\documentclass{article}
\usepackage{iclr2027_conference,times}

\usepackage{amsmath,amsfonts,bm}

\def\eqref#1{equation~\ref{#1}}
\def\1{\bm{1}}

\DeclareMathAlphabet{\mathsfit}{\encodingdefault}{\sfdefault}{m}{sl}
\SetMathAlphabet{\mathsfit}{bold}{\encodingdefault}{\sfdefault}{bx}{n}

\usepackage{hyperref}
\usepackage{url}
\usepackage{graphicx}
\usepackage{wrapfig}
\usepackage{subcaption}
\usepackage{mathtools}
\usepackage[most]{tcolorbox}
\usepackage{booktabs}
\usepackage{cleveref}

\graphicspath{ {graphics/} }

\newcommand{\ie}{\textit{i}.\textit{e}.}
\newcommand{\eg}{\textit{e.g.}}

\newtcolorbox{promptbox}[2][]{
  enhanced,
  colback=gray!8,
  colframe=gray!60,
  boxrule=0.4pt,
  arc=2mm,
  left=3mm,right=3mm,top=2mm,bottom=2mm,
  title=#2,
  #1
}

\definecolor{tmplframe}{HTML}{1B4F5C}
\definecolor{tmplback}{HTML}{F4F8F9}

\newcommand{\slot}[1]{\texttt{\footnotesize\{#1\}}}

\newenvironment{tmpllist}{%
    \begin{itemize}\footnotesize
    \setlength{\itemsep}{1.5pt}\setlength{\parskip}{0pt}
    \setlength{\topsep}{2pt}\setlength{\leftmargini}{1.2em}}%
    {\end{itemize}}

\newtcolorbox[
    auto counter,
    number within=section,
    crefname={Box}{Boxes},
    Crefname={Box}{Boxes},
    list inside=tmpl
]{templatebox}[2][]{%
    enhanced, breakable,
    colback=tmplback, colframe=tmplframe,
    boxrule=0.6pt, arc=1.5pt,
    left=6pt, right=6pt, top=5pt, bottom=5pt,
    toptitle=5pt, bottomtitle=5pt,
    fonttitle=\bfseries\small,
    title={Box~\thetcbcounter: #2}, #1}

\title{Latent Taxonomies: How Property Induction Develops in Language Model Pre-training}

\author{%
  Jan Sobotka$^{1,2,}$\thanks{Correspondence to: \texttt{jan.sobotka@epfl.ch}}
  \quad
  Ekaterina Goliakova$^{2,3}$
  \quad
  Marcin Detyniecki$^{2,4}$\\
  \textbf{Antoine Bosselut}$^{1}$
  \quad
  \textbf{Xavier Renard}$^{2,3}$
  \\[0.2em]
  $^1$EPFL \quad $^2$AXA  \quad $^3$Sorbonne University, CNRS, LIP6 \quad $^4$Polish Academy of Science, IBS PAN
}

\iclrfinalcopy
\begin{document}

\maketitle

\begin{abstract}
When told that robins have sesamoid bones, people are far more prone to extend the property to sparrows than to whales. This behavior, called category-based property induction, is well documented in humans, but whether and how it develops in language models trained only to predict the next token is unclear. To shed light on this, we construct a hierarchical taxonomy of synthetic entities grouped by the properties they share. We then inject facts from this latent taxonomy into the continued pre-training of Pythia and OLMo 2 models from 70M to 7B parameters, and track the development of inductive behaviors. We find that taxonomically aligned induction emerges from next-token prediction alone: a property learned for one entity transfers more strongly within its taxonomic category than across categories. Internally, entities that share a deeper category become more similar in activation space, and clustering recovers the full taxonomic tree from entity representations. Contrary to the usual expectation that scale helps, we find that induction weakens with model size when the data budget is fixed, suggesting that larger models memorize individual facts rather than compress them into shared structure. Induction also strengthens with more pre-training before the taxonomy injection, transfers to novel in-context properties, and can be overridden by a latent taxonomy implied by the prompt. Together, these results show that a model's inductive generalization reflects the latent organization of its training data, with implications for data curation and for predicting how far a learned fact will spread.
\end{abstract}

% \ificlrfinal \vspace{-6pt} \fi
\section{Introduction}
\label{sec:introduction}
% \ificlrfinal \vspace{-4pt} \fi
Human semantic cognition is organized around categories composed of related entities~\citep{rogers2004semantic}. We do not store the properties of every object independently; instead, we infer them from category membership. This lets us predict unobserved features of novel or partially observed entities \citep{rogers2004semantic}. Such an inference is called \emph{category-based property induction}\footnote{We refer to this as \emph{property induction} or just \emph{induction} in the rest of the paper.}: learning that a novel property holds for one entity, we extend it more strongly to taxonomically close entities than to distant ones. For example, when told that robins have sesamoid bones, we expect sparrows to have them too, but are far less sure about whales \citep{osherson1990induction}. This graded generalization is not innate, but develops through increasingly systematic induction across human development \citep{keil1979semantic,mandler1993concept}.

Knowing whether language models (LMs) acquire comparable structure matters for human-AI collaboration and for anticipating LM outputs. In particular, useful communication presupposes shared categories: a chatbot assistant is only reliable if its notion of which things are alike matches its users'. Additionally, inductive generalization of properties is the mechanism by which a fact stated about one entity influences predictions about others, so understanding it is a prerequisite for anticipating what a model will infer, and what it will overgeneralize, from its data. Such understanding is particularly relevant for training data curation and AI safety~\citep{anwar2024foundational}.

Prior work on property induction in LMs focuses primarily on the final behavior of pre-trained models \citep{han2024inductive,rodriguez2025characterizing}, but little is known about how such induction develops. Work that studies models during training fine-tunes for yes/no property judgment \citep{misra2022propertyinduction}, which differs from the standard next-token objective under which LMs acquire most of their semantic knowledge during pre-training~\citep{zhou2023lima,gekhman2024fine}. Studies that evaluate pre-trained models on real entities cannot control entity-property co-occurrences the models have seen \citep{misra2022propertyinduction,han2024inductive,rodriguez2025characterizing}, and do not have full control over the ground-truth taxonomy. We close these gaps with a synthetic but realistic setup. Specifically, we construct a controlled taxonomy of novel entities, express it as natural-language text, and inject it into the continued pre-training of real LMs ranging from 70M to 7B parameters alongside their original pre-training corpora. Because every entity and property is a pseudoword, co-occurrence is fully controlled, and leakage is highly unlikely by construction, while the training signal remains ordinary next-token prediction on realistic text.

Using this controlled setup, we make the following findings:
\begin{enumerate}
\vspace{-2pt}
    \item We show that LMs develop taxonomically aligned inductive generalization purely from next-token prediction, supported by a latent taxonomy in the training data (\autoref{sec:emerge}).
    \item This generalization is reflected in the model's internal representations: entity representations become more similar the deeper their shared category, and the full taxonomy can be recovered from them via clustering, validating \citet{saxe2019}'s theory (\autoref{sec:reps}).
    \item We find that the strength of acquired induction grows with the number of pre-training steps on taxonomy-unrelated text. Counterintuitively, given the same data budget, increasing model size \emph{reduces} correct induction (\autoref{sec:scale}).
    \item We show that learning a conflicting taxonomy on top of an old one erases the old grouping, and the new one is learned less well than without the prior commitment (\autoref{sec:flexibility}).
    \item We find that distilling only a latent taxonomy into a model is enough to enable in-context inductive generalization over novel data, and when in-context and in-weights taxonomies disagree, the model generalizes according to the context (\autoref{sec:context}).
\end{enumerate}

\vspace{-4pt}
\section{Experimental setup}
\label{sec:experimental_setup}
\vspace{-2pt}
\subsection{Models}
\label{subsec:models}

We validate our findings across two model families and a range of sizes. The most important hyperparameters and settings used for their continued pre-training are reported in \autoref{tab:pretraining_hyperparameters} in the Appendix.

\textbf{Pythia.}
We use the Pythia suite \citep{biderman2023pythia} of decoder-only transformers trained with an identical data order on the Pile \citep{pile}, with public intermediate checkpoints spanning the 143k-step training run. The shared data order and dense checkpointing make Pythia well suited to developmental analysis. We continue pre-training from a released checkpoint, resuming the original data stream while mixing in our synthetic taxonomy (\autoref{subsec:training_data}). Unless stated otherwise, we resume the optimizer state and start from the 50k/143k checkpoint of the 160M model. This choice allows us to study realistic intermediate stages of pre-training when naturally occurring taxonomic data could be seen by the model for the first time (\eg, in multi-epoch training).

\textbf{OLMo~2.}
We complement Pythia with 1B and 7B OLMo~2 \citep{olmo2}, trained for 1.9M steps on OLMo-mix-1024. OLMo~2 differs from Pythia in architecture, tokenizer, and data distribution, so it tests whether our findings are family-specific. As with Pythia, we resume pre-training with the next-token objective from a released checkpoint (800k/1.9M).

\vspace{-2pt}
\subsection{Training data}
\label{subsec:training_data}
Studying induction \emph{as it develops} requires two things: a data distribution close to real pre-training, and complete control over which entities co-occur with which properties. Real corpora provide the first but not the second; purely synthetic training provides the second but not the first. We combine them by mixing a controlled synthetic taxonomy into an otherwise unchanged pre-training stream.

\begin{figure}[t]
    \centering 
    \includegraphics[width=\linewidth]{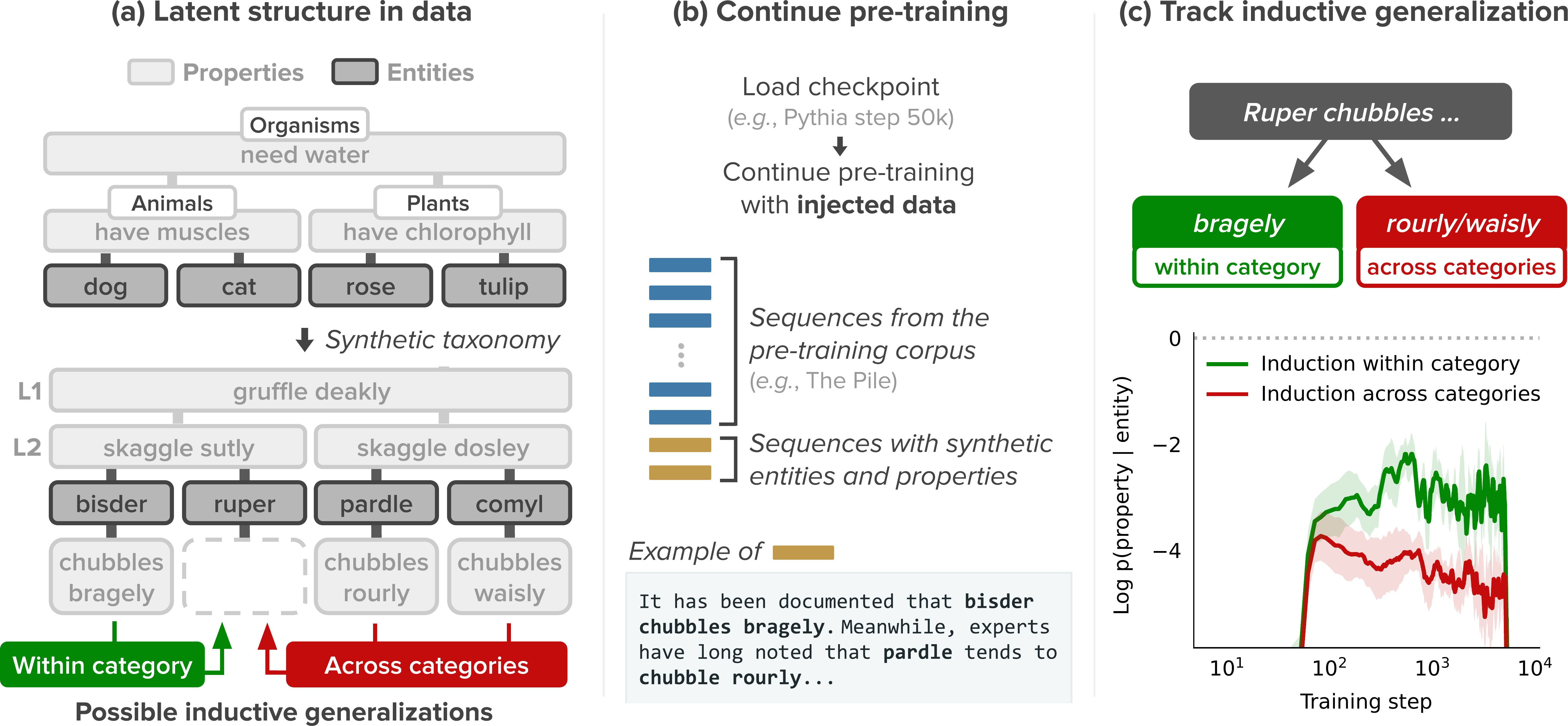}
    \vspace{-16pt}
    \caption{Overview. (a) Example natural (top) and synthetic (bottom) taxonomies of depth $D=2$; experiments use $D=3$. (b) We continue pre-training Pythia and OLMo 2 with synthetic sequences mixed in, and (c) track inductive generalization (Pythia 160M shown; see \autoref{fig:main}).}
    \label{fig:taxonomy}
\end{figure}

\textbf{Synthetic taxonomy of concepts.}
We define a balanced binary tree of depth $D$ over $2^D$ leaf entities. Each internal node carries \emph{shared properties}, and an entity inherits the properties of its ancestors, so two entities that share a deeper node share more properties. Such a hierarchy may, for example, reflect a taxonomy of organisms (\autoref{fig:taxonomy}a, top): animals and plants are the two children of the root, leaves are instances (\eg, dog, cat, rose, tulip), and properties are features shared at each level (\eg, animals have muscles), so that dog and cat share more properties than either does with rose.

The taxonomy is thus defined implicitly (latently) by which entities share which properties. To test whether LMs trained on these shared properties discover this latent taxonomic structure, we evaluate whether their inductive generalization becomes taxonomically aligned. In particular, we add a probe property $P$ that is seen during training for only one entity and held out from its sibling in the same lowest-level category (the \textit{target} entity; \eg, \textit{bisder chubbles bragely} is seen and \textit{ruper} is the target in \autoref{fig:taxonomy}a, bottom). An increased probability of $P$ for the target then measures within-category inductive generalization. For comparison to across-category induction, we also add control properties $C_i \in C$, each seen during training with exactly one entity $i$ outside the target's lowest-level category. We describe how we score these quantities in \autoref{subsec:evaluation}.

We use pseudoword entity names and properties produced by the Wuggy generator~\citep{keuleers2010}, which derives pronounceable non-words from English. Pseudowords guarantee that the model has no prior association for any entity or property, giving us full control over which entities co-occur with which properties, and how they are organized in the ground-truth taxonomy. Unless stated otherwise, we set $D = 3$ in all experiments reported in this work, giving 8 leaf entities, and each internal node carries two shared properties. We provide an example taxonomy in \autoref{fig:taxonomy}a. For training, we render the taxonomy into text by stating the pairs of entities and properties in natural-language sentences sampled uniformly at random from a pool of 30 templates (Appendix, Box~\ref{box:train_templates}). Further details, including tokenization of pseudowords, can be found in \autoref{app:setup}.

\textbf{Data mixing.}
Each training batch contains 32 sequences of 2,048 (Pythia) or 4,096 (OLMo 2) tokens. For the first 5,000 of the total 10,000 steps of continued pre-training, we inject synthetic facts from our taxonomy into each training batch (\autoref{fig:taxonomy}b). In particular, we inject (1) one sequence per depth level in the taxonomic tree, where each sequence contains all entities and properties at that level, and (2) one sequence with the probe and control properties. We randomize the order of sentences with the entity-property facts within these sequences for each batch, so models cannot rely on positional cues. For a balanced taxonomic tree of depth $D=3$, this methodology results in training batches with 28 sequences from the background pre-training corpus and 4 sequences with synthetic facts. We ensure that the frequency of all entity-property pairs at a particular level is matched, thus balancing exposure to each part of the latent taxonomic tree. Sequences shorter than the context length are padded with a padding token that we exclude when calculating the loss.

These injections ensure that the synthetic signal is a relatively small share of tokens ($\approx$2--4\% after padding), as a rare but consistent fact would be in real data. While the resulting frequency is not calibrated to any measured rate in natural corpora, we believe this design is a good initial step toward approximating real-world pre-training while still allowing controlled studies of how data organization shapes generalization. We provide a comparison to training with lower ratios of synthetic data and the original batch size in \autoref{app:batch_size} in the Appendix.

\subsection{Evaluation}
\label{subsec:evaluation}
We call induction \emph{correct} when it follows the latent taxonomy: a novel property observed for one entity should transfer to entities that share its deepest category, and less to more distant ones. For a target entity and a prefix template, we score $\log p(\text{property}\mid\text{entity})$\footnote{For readability, we write $\log p(\text{property}\mid\text{entity})$ throughout, omitting the prompt template and the surrounding text that the probability is also conditioned on.} for several candidate completions and average over a set of evaluation templates (Appendix, Box~\ref{box:eval_templates}). We track three quantities:
\begin{itemize}
    \item \emph{Memorization}: the probability of a probe property $P$ for an entity that was trained with it (\eg, \emph{bisder chubbles bragely} in \autoref{fig:taxonomy}a, bottom).
    \item \emph{Induction within category}: the probability of a probe property $P$ for a same-category sibling that was \emph{not} trained with it (\eg, \emph{ruper chubbles bragely} in \autoref{fig:taxonomy}a, bottom).
    \item \emph{Induction across categories}: the average probability of a control property $C_i \in C$ for the same target entity as the induction-within-category measurement, where the $C_i$ belongs to a different category (\eg, \emph{ruper chubbles rourly} in \autoref{fig:taxonomy}a, bottom).
\end{itemize}

Correct taxonomic induction means within-category $>$ across-category (\autoref{fig:taxonomy}c). We summarize the gap as the \emph{induction advantage}, the difference in log-probability between within- and across-category induction:
\vspace{-2pt}
\begin{equation}\label{eq:induction_advantage}
    \text{induction advantage}
    \coloneq
    \frac{1}{|F|}
    \sum_{f \in F}
    \bigg(
    \log p\big(P \mid f(E)\big) -
    \frac{1}{|C|}
    \sum_{C_i \in C} \log p\big(C_i \mid f(E)\big)
    \bigg),
\end{equation}
\vspace{-2pt}
where $E$ refers to the target entity and $f(E)$ denotes the prompt template $f \in F$ filled with~$E$.

\section{Results}
\label{sec:results}
To make sure that our results do not depend on any particular pseudoword vocabulary or taxonomy, we repeat each experiment across three random seeds (unless stated otherwise), each of which regenerates the entire pseudoword vocabulary and reassigns it to the taxonomic tree.

\begin{figure}[t]
    \centering 
    \includegraphics[width=\linewidth]{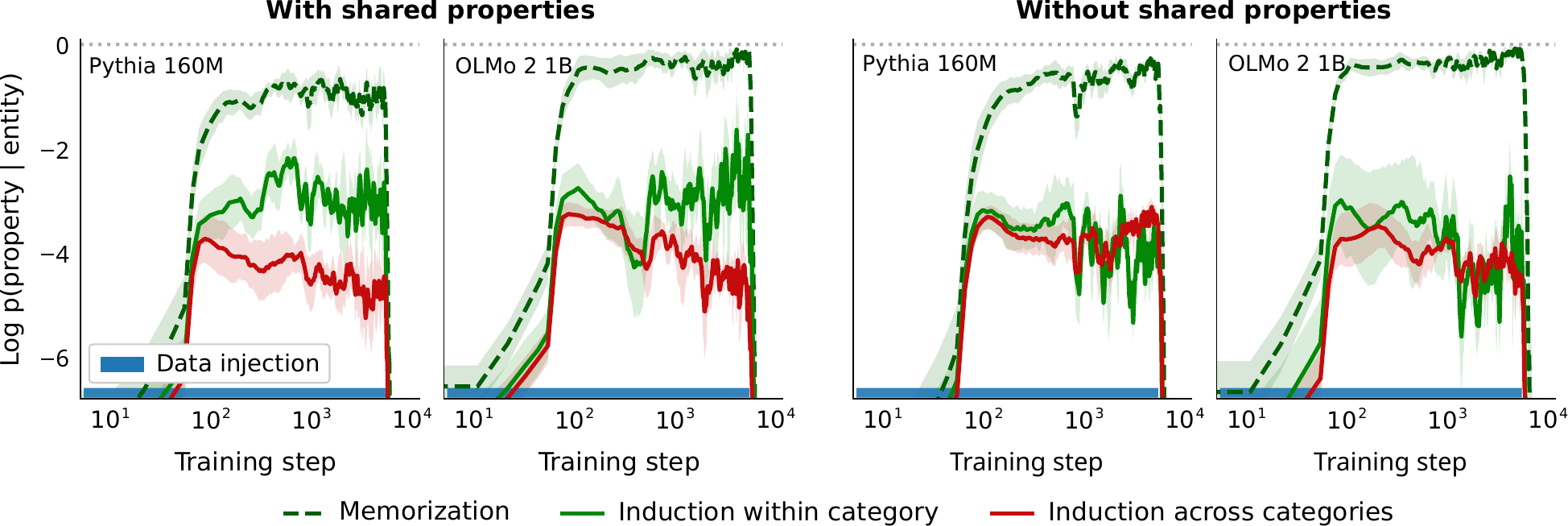}
    \vspace{-14pt}
    \caption{$\text{Log}\ p(\text{property}\mid  \text{entity})$ over continued pre-training of Pythia~160M and OLMo~2~1B. With shared properties (left), memorization (dashed) $>$ induction within category (green) $>$ induction across categories (red). Without shared properties (right), within- and across-category induction collapse onto each other. Shaded regions: standard error of the mean across three random seeds.}
    \label{fig:main}
\end{figure}

\subsection{LMs inductively generalize along the latent taxonomy}\label{sec:emerge}
The next-token objective does not ask a model to build categories explicitly: the taxonomy is only implicit (latent) in which entities share which properties. We therefore begin by asking whether taxonomically aligned induction appears at all under ordinary pre-training, or whether it requires the specialized supervision used in earlier work~\citep{misra2022propertyinduction}.

To answer this question, we track the memorization and induction within and across categories over continued pre-training of Pythia~160M and OLMo~2~1B. As \autoref{fig:main} (left) shows, all three quantities rise as the taxonomy is learned, but they separate from each other as training progresses: memorization is highest, induction within category sits well above induction across categories (\ie, \emph{correct induction}), and the ordering persists for the rest of training.

The correct generalization along the latent taxonomy is, however, forgotten once the evidence for the taxonomy is removed from the training signal  (Appendix, \autoref{fig:forgetting}). Taxonomic structure is therefore maintained only while the data continues to support it. \citet{chang2024knowledge} report similar catastrophic forgetting for individual injected facts during pre-training, and our result extends it from the recall of single facts to the taxonomic organization that supports generalizing them.

\textbf{Ablation: no shared properties.}\label{sec:ablation}
When we remove the shared properties from training, effectively leaving no signal about which entities belong together in a category, within- and across-category induction collapse onto each other (\autoref{fig:main}, right). This further demonstrates that the correct induction is driven by the model learning the latent taxonomy.

\begin{figure}[t]
    \centering
    \includegraphics[width=\linewidth]{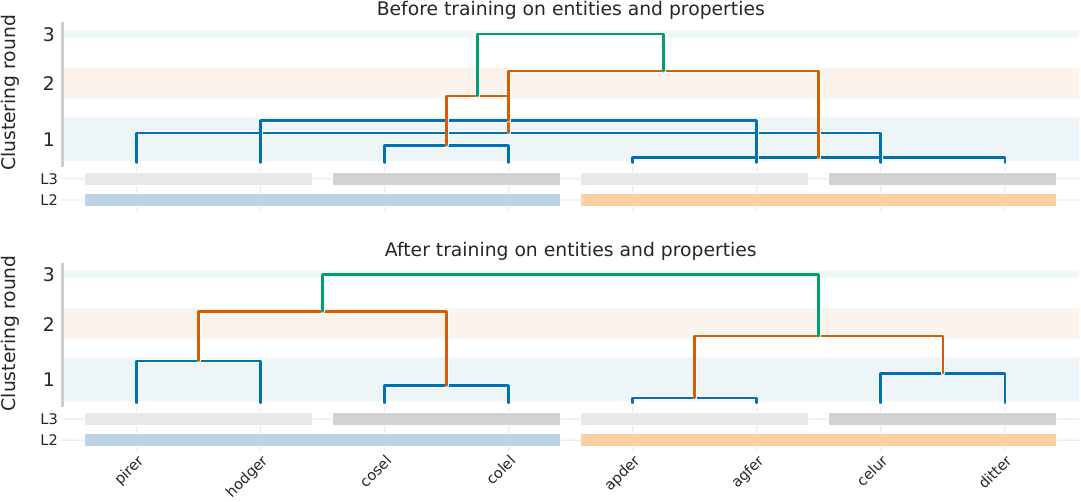}
    \vspace{-16pt}
    \caption{Hierarchical recovery of the taxonomy from entity representations of OLMo~2~1B. Dendrograms before vs. after continued pre-training on entities and properties. Bars below each dendrogram mark the ground-truth L3 and L2 categories (clustering details in Appendix, \autoref{app:clustering}).}
    \label{fig:cluster_olmo}
    \vspace{-2pt} 
\end{figure}

\subsection{The taxonomy is recoverable from entity representations}
\label{sec:reps}

\textit{Where does the correct induction come from?} \citet{saxe2019} showed that in deep linear networks, hierarchically organized concepts that share deeper categories become more similar in representation. Whether this carries over to Transformer-based LMs is not obvious, since they compose many non-linear operations, use attention rather than fixed weights between fixed units, and are trained on natural text rather than on clean one-hot attribute vectors. We test a stronger case of \citet{saxe2019}'s theory: \textit{Can the entire tree, including its branching structure, be fully reconstructed from the representations?} Using a standard clustering technique, we find that it can.

In particular, we cluster the last-token entity representations from mid-to-late layers of our trained checkpoints (layer 8 out of 12 for Pythia 160M, and layer 8 out of 16 for OLMo 2 1B), following prior work that finds these intermediate layers carry more high-level semantic information than early or final ones~\citep{nostalgebraist2020interpreting,geva2021transformer,skean2025layer}. We use \emph{Matching Affinity Clustering} (MAC)~\citep{hajiaghayi2021improved} with cosine distance for this purpose (details in Appendix, \autoref{app:clustering}).

After the data injection phase of our continued pre-training, we find that the recovered tree matches the ground-truth taxonomy perfectly for OLMo~2~1B (\autoref{fig:cluster_olmo}, bottom) and near-perfectly for Pythia~160M (Appendix, \autoref{fig:cluster_pythia}, bottom). The same models before taxonomy injection do not recover it (\autoref{fig:cluster_olmo} and \autoref{fig:cluster_pythia}, top), so the structure is not an artifact of the clustering algorithm, the tokenization, or the geometry of the layer we selected. Additionally, \autoref{app:repsim} in the Appendix shows that the average similarity between entity representations grows with the depth of their shared category, providing further evidence that the representations reflect the taxonomy.

\subsection{Development and the effect of scale}
\label{sec:scale}
The human developmental literature suggests that induction in later learning stages might be more taxonomy-aligned, since children generalize more systematically as their existing category knowledge matures \citep{fisher2005induction,badger2012evidence}. \textit{Does the same hold for language models?} Additionally, in terms of model scale, the usual expectation in deep learning is that bigger is better, since loss falls predictably with parameters, data, and compute \citep{kaplan2020scaling,hoffmann2022training}, and downstream capability generally follows \citep{brown2020language,wei2022emergent}. Whether the same scaling trends hold for inductive generalization is an open question.

\textbf{More mature checkpoints acquire induction more strongly.}
We repeat the experiment from \autoref{fig:main} for Pythia~160M checkpoints spanning 0k to 143k pre-training steps, and for OLMo 2 1B checkpoints from 0k to 1.9M steps. Specifically, we vary the base checkpoint from which we continue pre-training, holding the taxonomy, vocabularies, and number of training steps the same. To quantify the strength of inductive generalization with a single number, we average the induction advantage~(\autoref{eq:induction_advantage}) over the last 1,000 continued pre-training steps of the data injection. 

\begin{figure}[t]
    \centering
    \includegraphics[width=0.98\linewidth]{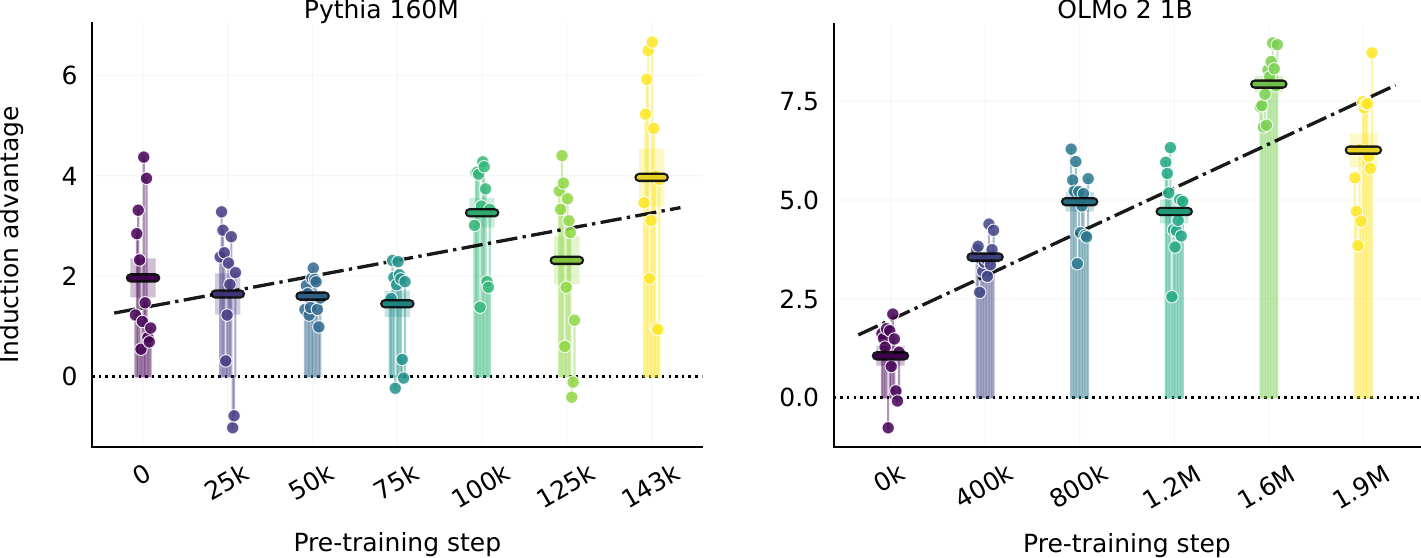}
    \vspace{-4pt}
    \caption{More mature checkpoints acquire induction more strongly. Shown is the induction advantage averaged over the last 1,000 steps of synthetic data injection in continued pre-training from various checkpoints (x-axis). Circles show induction advantage for individual evaluation templates, while horizontal lines show the average. All reported values are averages from three random seeds.}
    \label{fig:checkpoints}
\end{figure}

As shown in \autoref{fig:checkpoints}, the average induction advantage increases with the base checkpoint's pre-training step. With the exception of a few checkpoints, the overall trend shows that a model that has learned more general language before the taxonomy is introduced learns to generalize over it more strongly. The capacity for correct generalization is, therefore, something that pre-training facilitates.

\begin{figure}[t]
    \centering
    \includegraphics[width=0.84\linewidth]{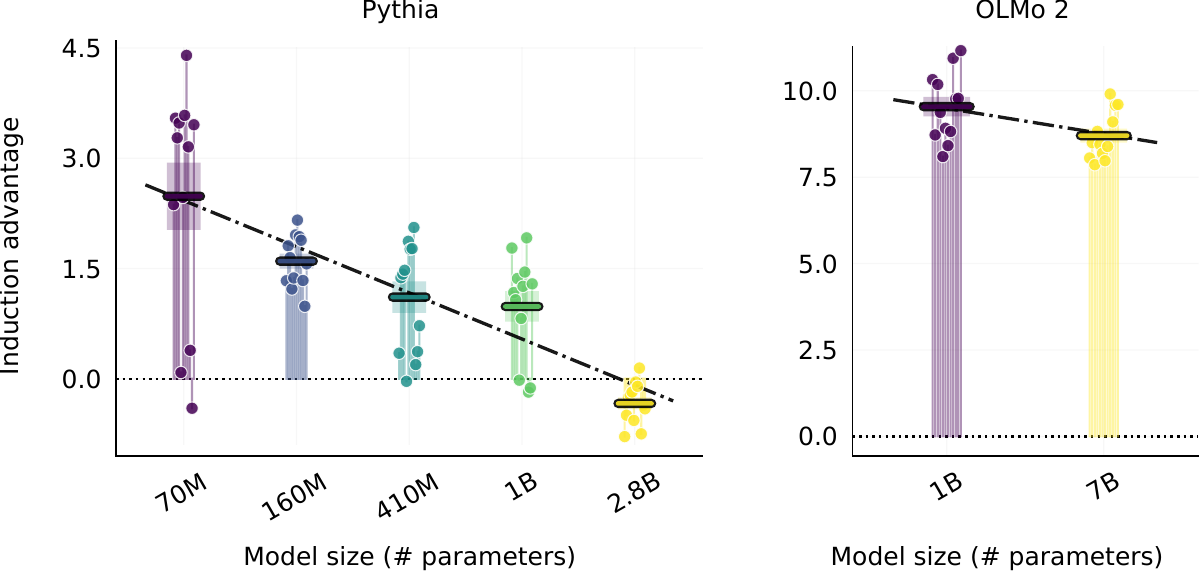}
    \vspace{-4pt}
    \caption{At a fixed data budget, scale hurts. Shown is the induction advantage averaged over the last 1,000 data injection steps vs.\ model size. Circles indicate induction advantage for individual evaluation templates, while horizontal lines show the average induction advantage. All reported values are averages from three random seeds.}
    \label{fig:scale}
\end{figure}

\textbf{At a fixed data budget, scale hurts.}
Scaling the model has the opposite effect. Across Pythia 70M--2.8B and OLMo 2 1B--7B trained on identical injected data, the induction advantage \emph{decreases} with model size (\autoref{fig:scale}). The 70M Pythia model and OLMo 2 1B show the strongest correct induction, while larger models exhibit a relatively weak one, most clearly at the endpoints.

We hypothesize that this reflects a trade-off between memorization and generalization. Larger models may fit the injected facts fast enough to memorize them directly, which would weaken the pressure to generalize along the taxonomy to minimize loss on the shared properties. A smaller model, in contrast, may be unable to store every entity and property pair separately fast enough, making it cheaper to represent the entities in terms of the categories they share, which in turn would produce correct induction.

If this hypothesis holds, larger models should memorize the injected facts faster, but generalize along the taxonomy more slowly. We test this prediction on the Pythia family, which provides a denser span of model scales. In \autoref{fig:scale_speed} in the Appendix, we plot memorization as defined in \autoref{subsec:evaluation} over training for each model size (left), along with the induction advantage averaged over the first 1,000 continued pre-training steps (right). The ordering of the curves on the left confirms that larger models memorize faster, while the downward trend on the right shows that they inductively generalize along the taxonomy more slowly.

These two findings go in opposite directions and complicate a common assumption: more pre-training improves property induction, but more parameters at a fixed data budget do not. Practitioners injecting new domain knowledge into larger models should therefore expect them to more readily recall rather than inductively generalize, unless the data gives the taxonomy enough support to make generalizing worthwhile.

\subsection{Flexibility and interference across taxonomies}
\label{sec:flexibility}
Category structure in the real world is not always fixed: fields reorganize, companies restructure, and a model will eventually need to revise a taxonomy it has already learned. To study how models handle such revision, we first train LMs on taxonomy $T_1$, and then continue training on a conflicting taxonomy $T_2$ that reuses the same entity names but regroups them into different categories (schematic of the experiment in \autoref{fig:conflict_setup} in the Appendix).

The results in \autoref{fig:conflict} demonstrate two things. First, correct induction under $T_1$ is gradually forgotten once $T_2$ training begins (dashed lines on the left). Second, the correct induction within the new taxonomy $T_2$ is weaker than the induction a fresh model would acquire for the same taxonomy (\autoref{fig:conflict}, right). We interpret this as interference: a prior taxonomic commitment makes a conflicting one harder to learn, because the entities carry a residue of their earlier categorization that the new training has to work against.

\begin{figure}[t!]
    \centering
    \includegraphics[width=\linewidth]{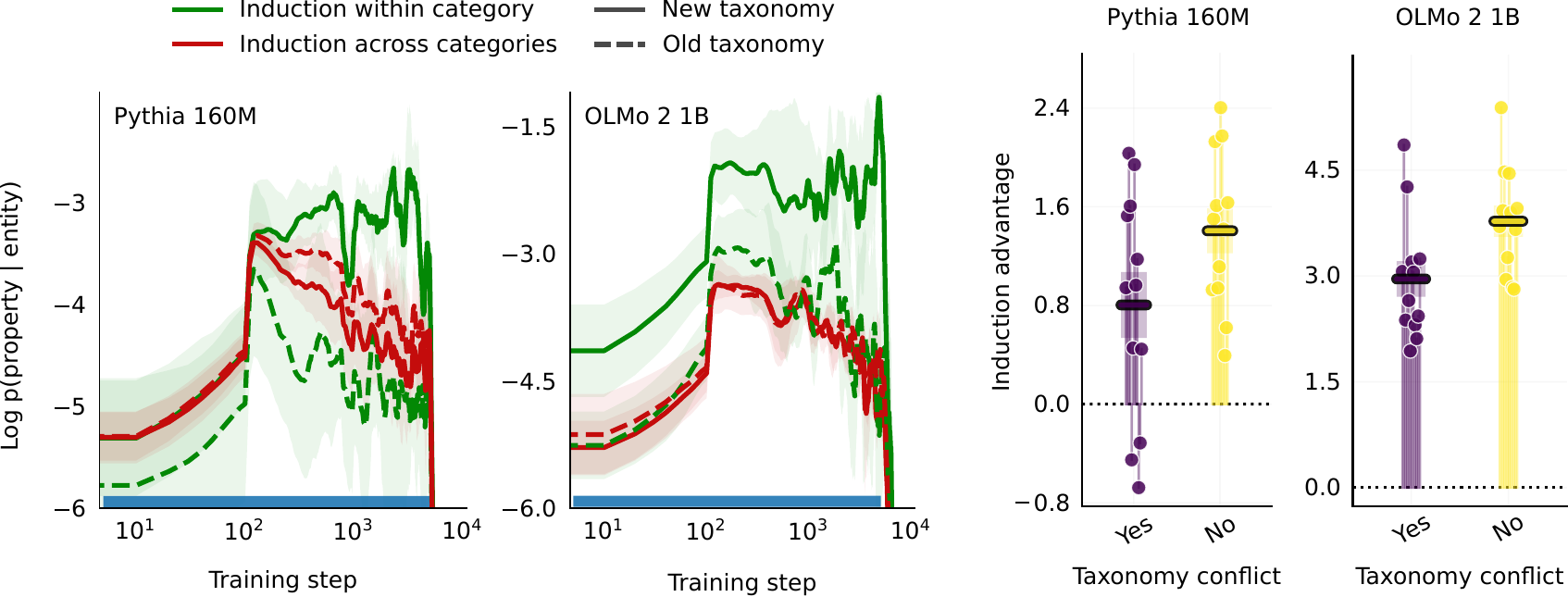}
    
    \vspace{-6pt}
    \caption{Left: induction within vs.\ across categories during two-phase training (Taxonomy conflict: \emph{Yes}), measured against both the old taxonomy $T_1$ (dashed) and the new taxonomy $T_2$ (solid). Right: induction advantage reached under $T_2$ when the model was first trained on a conflicting taxonomy (\emph{Yes}) vs.\ trained on $T_2$ from the same base checkpoint without the prior phase (\emph{No}).}
    \label{fig:conflict}
\end{figure}

\subsection{In-weights taxonomy shapes in-context induction}
\label{sec:context}

All results presented so far concern knowledge the model acquired through training. In real-world deployment, however, information a model is given frequently arrives only in the prompt. Training and in-context learning (ICL) therefore meet in every real interaction, and how they combine determines what a user can actually get a model to do. We pose two research questions: (1) \textit{Does a taxonomy learned in the weights help the model reason about properties it only sees in context?} and (2) \textit{What happens when the prompt and the weights disagree?}

\begin{figure}[t]
    \centering
	\includegraphics[width=\linewidth]{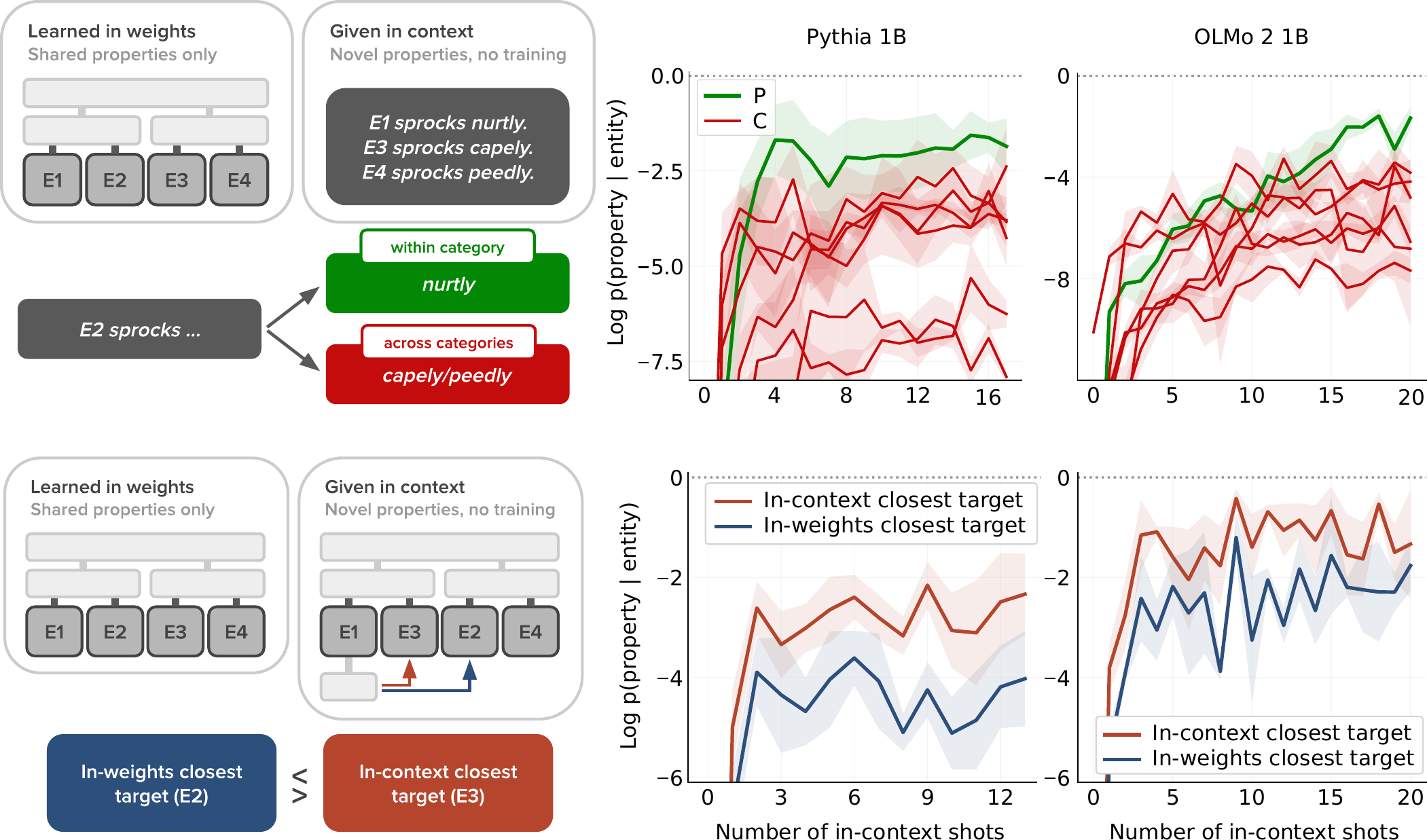}
    \vspace{-12pt}
    \caption{In-weights taxonomy shapes in-context induction. Left: schematics of the two setups; models are trained only on shared properties, so probe and control properties appear solely in the prompt. Top: inductive generalization of the probe $P$ (green) vs.\ each control $C_i$ (red) to target E2; one shot lists the probe and all control properties. Bottom: generalization of $P$ to the in-context closest target E3 vs.\ the in-weights closest target E2; one shot lists all properties in the in-context taxonomy. Shots are capped by each model's context window; sentences use randomly sampled evaluation templates (Box~\ref{box:eval_templates}); shaded regions show SEM over five random seeds.}
    \label{fig:incontext}
\end{figure}

\textbf{Learning a taxonomy enables in-context induction of novel properties.}
We train Pythia~1B\footnote{We use Pythia~1B rather than the 160M model because we found the smaller model unable to copy a property from the context onto a repeated entity, which is a prerequisite for the in-context form of this task.} and OLMo~2~1B only on the shared properties of the taxonomy, so they learn the category structure but never see the probe or control properties. We find that a few in-context examples that assign properties to entities are then enough for the models to generalize inductively. In particular, as we repeat the set of sentences containing the probe and control properties (\emph{in-context shot}), the within-category generalization of the probe property $P$ is preferred over the control properties $C_i$ (\autoref{fig:incontext}, top). This shows that acquiring taxonomic structure in the weights enables models to apply it to novel in-context properties.

\textbf{Context overrides weights under conflicting taxonomies.}
Users may supply categorizations that contradict pre-training: a document seen in pre-training might organize products by supply chain, while a user groups them in the prompt by function. Whether the model follows such a correction or reverts to its learned taxonomy sets a limit on how far it can be steered by context.

We test this conflict by first training Pythia~1B and OLMo~2~1B so that their weights favor stronger inductive generalization of the probe property $P$ from entity E1 to entity E2 (\autoref{fig:incontext}, \emph{in-weights closest target}) than to entity E3 (\emph{in-context closest target}). Then we evaluate these trained models with in-context shots supporting the opposite grouping (stronger E1 $\rightarrow$ E3 generalization than E1 $\rightarrow$ E2). We can see that as in-context shots accumulate, inductive generalization to the in-context closest target E3 overtakes generalization to the in-weights target E2, and the model generalizes according to the context taxonomy (\autoref{fig:incontext}, bottom).

This result demonstrates that a handful of in-context examples can override a taxonomy learned in pre-training from far more evidence. One potential implication of this finding is that when provided with a misleading context, a model can be steered in the wrong direction even if the pre-trained categorization was correct. On the other hand, it appears that a user can rewrite a taxonomy at the prompt level when the one from training becomes outdated.

\section{Related work}
\label{sec:related_work}
\textbf{Property induction in language models.} 
\citet{misra2022propertyinduction} fine-tune LMs to answer yes/no questions about propositions and find that inductive preferences track taxonomic information latent in the data. \citet{rodriguez2025characterizing} show that property inferences of instruction-tuned LMs are jointly explained by taxonomic relations and pairwise categorical similarity, and \citet{han2024inductive} find broadly human-like patterns in GPT-3.5 and GPT-4, evaluating purely in-context with premises in the prompt. All three read out induction from a final checkpoint after task-specific supervision or from the prompt alone, and all use real entities. We instead study in-weights induction acquired under the next-token objective alone, track it across pre-training rather than at an endpoint, and use pseudowords for both properties and entities, which removes the taxonomic structure that prior knowledge of the entities would otherwise supply. \autoref{app:cogsci} reviews the cognitive-science literature that defines this paradigm and motivates some of the experimental choices.

\textbf{Learning dynamics and representational geometry.} 
Our developmental analysis is anchored in the connectionist theory of semantic development \citep{rogers2004semantic,saxe2019}, in which linear networks trained on hierarchical entity-property matrices admit closed-form learning trajectories predicting coarse-to-fine differentiation of representations, mirroring conceptual development in children \citep{keil1979semantic,mandler1993concept}. We test whether this survives in deep Transformer-based LMs trained on natural text, using open intermediate checkpoints \citep{biderman2023pythia}. This complements work on factual acquisition during pre-training \citep{chang2024knowledge,zucchet2025facts}, which treats facts atomically rather than as inductively related category members. Relatedly, \citet{park2025geometry} analyze how hierarchical concepts are encoded as linear structure \citep{park2024linear}, and \citet{nava2026hierarchical} argue this geometry emerges from co-occurrence statistics; both target end-of-pre-training checkpoints on natural concepts, whereas our controlled taxonomy lets us track how such representations develop.

\section{Conclusion}
\label{sec:conclusion}
We study whether category-based property induction, a behavior long documented in humans, develops in language models trained only to predict the next token. By injecting a controlled synthetic taxonomy\footnote{We discuss the limitations of our setup in \autoref{app:limitations}.} of pseudoword entities and properties into the continued pre-training of Pythia and OLMo 2, we show that taxonomically aligned induction emerges from the next-token objective alone, and that it is written into the model's representations. Specifically, entities grouped under a deeper category move closer together in activation space, and the full taxonomic tree can be recovered from those representations by clustering. We further find that induction strengthens with the amount of pre-training preceding the injection, but weakens with model size at a fixed data budget. Our analysis of learning speed supports one explanation: larger models memorize each fact faster, possibly reducing the pressure to compress entities into the categories they share. Beyond acquisition, we find that a taxonomy in the weights interferes with learning a conflicting one, yet transfers to novel properties supplied in context and can be overridden by a taxonomy implied by the prompt. Together, these results show that a model's inductive behavior is a property of the latent organization of its training data, that strong recall and strong inductive generalization need not coincide, and that the resulting structure is more easily changed by context than rewritten by further training.

\subsection*{AI use statement}
In this work, we used generative AI tools for providing feedback on research methodology and implementing methods.
We have not used generative AI tools for proposing or refining hypotheses, generating synthetic datasets or cleaning and reformatting existing datasets, helping to develop theoretical models or conceptual frameworks,  supporting qualitative and thematic data analysis, or interpreting the results. Formulating mathematical claims, providing critical ingredients for proving mathematical claims, assisting in the writing of proofs, and assisting with translation are not applicable to this work.
Additionally, we used generative AI tools for creating or editing software code, drafting parts of a research paper, editing a research paper to improve readability, and identifying relevant literature. We have reviewed all AI-assisted work. In particular, LLM-generated code was verified and tested for correctness; all parts of the paper were ultimately written by the authors; and relevant literature identified by LLMs was used only for the authors to review and subsequently discuss in the paper. We take responsibility for the final content of this work, including text, claims, or artifacts produced with the aid of generative AI.

\subsection*{Reproducibility statement}
All components needed to reproduce our results are described in the paper and its appendix. The models and released checkpoints we continue pre-training from are specified in \autoref{subsec:models}, and the full set of optimization and batch-composition hyperparameters is given in Table~\ref{tab:pretraining_hyperparameters}, together with how the data stream and learning rates are resumed from the released checkpoints (\autoref{app:hyperparams}). The construction of the synthetic taxonomy and the regeneration of the pseudoword vocabulary across random seeds are described in \autoref{subsec:training_data} and \autoref{app:taxonomy}. The complete pools of training and evaluation templates are listed in Box~\ref{box:train_templates} and Box~\ref{box:eval_templates}, respectively. Subsection~\ref{subsec:training_data} specifies how the synthetic data is mixed into each training batch, and \autoref{app:batch_size} reports the same experiment at the original pre-training batch size. The taxonomy-conflict and in-context experiments are detailed in \autoref{app:conflict_setup} and \autoref{sec:context}, respectively. On the evaluation side, the scored quantities are defined in \autoref{subsec:evaluation} and Equation~\ref{eq:induction_advantage}, the constraint that makes completions comparable in \autoref{app:completion_space}, the tokenization of pseudowords in \autoref{app:tokenization}, the extraction of entity representations in \autoref{sec:reps}, the clustering procedure used to recover the taxonomy in \autoref{app:clustering}, and the smoothing of training curves in \autoref{app:presentation}. Every reported number is averaged over three random seeds (five for the in-context experiments; dendrograms show a single seed), each of which regenerates the entire vocabulary and reassigns it to the tree.% The code for our experiments and analysis will be made available upon publication of this work.

% \subsubsection*{Acknowledgments}
% Use unnumbered third level headings for the acknowledgments. All
% acknowledgments, including those to funding agencies, go at the end of the paper.

\bibliography{iclr2027_conference}
\bibliographystyle{iclr2027_conference}

\clearpage
\appendix
\section*{Appendix}
\section{Limitations}\label{app:limitations}
Our setup diverges from realistic settings in some aspects and leaves a few promising future directions open. We briefly describe each below.

First, the taxonomies we use are small, balanced, and noiseless, whereas real conceptual structures are often large, overlapping, admit exceptions, and depend on which property is being generalized. This is a deliberate trade-off: control over the ground-truth taxonomy and over every entity-property co-occurrence is what makes induction separable from recall, which is close to impossible when evaluating real entity-property pairs with pre-trained LLMs. We expect a larger or noisier taxonomy to weaken the measured effect rather than reverse it, since it depends only on entities sharing properties, and the theory this builds on predicts differentiation by shared depth for any hierarchically structured data~\citep{saxe2019}. Verifying this is left to future work.

Second, we inject four synthetic sequences into every training batch for 5,000 steps. While this is a small fraction of the tokens in each batch, appearing in every batch is a far denser exposure than the sparse, irregular recurrence of rare facts in real corpora. We chose this schedule because in preliminary experiments with sparser injection, the models failed to memorize even the entity-property facts they had been trained on, making any measurement of inductive generalization questionable (there is nothing to generalize from). The difficulty of acquiring rarely occurring facts, particularly for smaller models, is well documented \citep{kandpal2023large,chang2024knowledge,huang2026why}. At the same time, prior controlled studies of fact learning train from scratch on purely synthetic corpora \citep{allenzhu2024knowledge,zucchet2025facts}, which removes competition from natural text for model capacity and gradient signal completely. Our setup instead injects into an ongoing pre-training run on the model's original corpus, so the taxonomy is acquired alongside, and in competition with, real language. In this respect, our setting is a step closer to real pre-training than synthetic-only training.

Third, our scaling and checkpoint results vary one factor at a time. In particular, the finding that induction weakens with model size holds the injected data fixed, whereas in practice model size and data are scaled together \citep{hoffmann2022training}. Larger models may require proportionally more exposure before compressing entities into shared categories. Varying the data budget and model size jointly, and locating the exposure at which induction emerges for each size, is a promising avenue for future work.

\section{Category-based induction in cognitive science}
\label{app:cogsci}
 
The behavior we measure was defined and characterized in cognitive psychology, and that literature informs our definition of what counts as correct inductive generalization. \citet{rips1975inductive} first showed that inductive judgments track similarity rather than logical entailment. \citet{osherson1990induction} cataloged the robust human phenomena (premise typicality, diversity, similarity, monotonicity, and conclusion specificity) under a similarity-coverage account, and \citet{sloman1993feature} proposed a feature-based alternative requiring no explicit category representation; see \citet{heit2000properties,murphy2002big} for reviews. Later work casts induction as Bayesian inference over structured hypotheses such as taxonomic trees \citep{kemp2009induction,kemp2008discovery}. In parallel, the developmental literature established that the taxonomy supporting these inferences is itself learned, differentiating from coarse to fine over years of exposure \citep{keil1979semantic,mandler1993concept,pauen2002infancy}.
 
Two features of the standard experimental design in human studies are worth stating, because our setup with LMs imitates both. First, the inductively generalized property is a \emph{blank predicate}: a property that participants have no prior knowledge about, chosen so that judgments cannot come from what the reasoner already knows about the property and must instead come from the relation between the entities \citep{osherson1990induction}. Second, the categories and the underlying taxonomy are \emph{familiar} to the participants.

This second requirement is the constraint under which human studies operate. Because participants arrive with years of prior experience, an experimenter can guarantee no prior exposure to the property, but cannot do the same for the categories, which must be familiar for the task to make sense. The taxonomy participants actually use, therefore, has to be inferred by the experimenter, and their lifetime exposure to it cannot be quantified. Our setting removes this constraint by making both the properties and the entities novel, at the cost of a taxonomy far smaller and more regular than a human conceptual system. A second difference is the read-out: rather than collecting graded verbal judgments, we interpret token probabilities as argument strength.

\section{Experimental setup}
\label{app:setup}
 
\subsection{Hyperparameters}
\label{app:hyperparams}
\begin{table*}[t]
    \centering
    \renewcommand{\arraystretch}{1.16}
    \setlength{\tabcolsep}{8pt}
    \caption{Hyperparameters used for continued pre-training with injected synthetic taxonomy data. Learning rates are loaded from the optimizer checkpoints and are therefore slightly below the reported values due to their original (global) cosine schedules. A constant learning rate (LR) scheduler was chosen as it approximates the global cosine schedule well on our small training window of 10k training steps.}
    \vspace{0.15cm}
    \begin{tabular}{lcc}
        \toprule
        \textbf{Hyperparameter} & \textbf{Pythia} & \textbf{OLMo~2} \\
        \midrule
        \multicolumn{3}{@{}l}{\textbf{Optimization}}\\

        \textsc{Optimizer}                        & AdamW                  & AdamW\\
        \textsc{Learning rate}                    & $\approx 3 \times 10^{-4}$     & $\approx 4 \times 10^{-4}$\\
        \textsc{LR scheduler}                     & constant               & constant\\
        \textsc{Warmup steps}                     & 0                      & 0\\
        \textsc{Adam $\beta_1$}                   & 0.9                    & 0.9\\
        \textsc{Adam $\beta_2$}                   & 0.95                   & 0.95\\
        \textsc{Weight decay}                     & 0.1                & 0.1\\
        \textsc{Max gradient norm}                & 1.0                    & 1.0\\
        \textsc{Training steps}                   & 10,000                 & 10,000\\

        \multicolumn{3}{@{}l}{\textbf{Batch composition}}\\

        \textsc{Sequence length}                  & 2,048                  & 4,096\\
        \textsc{Batch size (sequences)} & 32            & 32\\
        \textsc{Batch size (tokens)} & 65,536           & 131,072\\

        \multicolumn{3}{@{}l}{\textbf{Precision and efficiency}}\\

        \textsc{Precision}                        & bfloat16               & bfloat16\\
        \textsc{Gradient checkpointing}           & no                     & no\\
        \textsc{Evaluation interval (steps)}         & 10                     & 10\\

        \bottomrule
    \end{tabular}
    \label{tab:pretraining_hyperparameters}
\end{table*}

For both Pythia and OLMo 2, the background corpus resumes from the position in the original data order that matches the loaded checkpoint, so the models continue with the sequences they would have seen next in their original runs. Learning rates are likewise loaded from the released optimizer checkpoints and are therefore slightly below the nominal values in \autoref{tab:pretraining_hyperparameters}, since the original runs used global cosine schedules. As a result, each model in the scaling experiments (\autoref{fig:scale} in \autoref{sec:scale}) keeps the learning rate of its own original run. During continued pre-training, we keep this learning rate constant, which approximates the global cosine schedule well over our short 10k-step window. The one exception is the comparison of base checkpoints from different pre-training steps (\autoref{fig:checkpoints} in \autoref{sec:scale}): there, we fix a single learning rate for all runs ($3 \times 10^{-4}$ for Pythia 160M and $4 \times 10^{-4}$ for OLMo 2 1B), so that differences between checkpoints are not driven by their position on the original learning-rate schedule.

\subsection{Construction of the synthetic taxonomy}
\label{app:taxonomy}

Every training sentence associates an entity with one of its properties (\eg, \emph{bisder gruffles deakly}), and the categories themselves are never explicitly stated. The taxonomy is therefore only latent in which entities share which properties. Our setup extends naturally to explicit category labels such as \emph{is an animal}: such a label is formally just another property shared by all entities under a node, so it is covered by the same construction.

The inheritance rule in our taxonomies fixes how much two entities have in common as a function of where they are in the taxonomic tree. With $D = 3$, every entity inherits six shared properties, two from each of the three levels above it, and any two entities agree on exactly as many of them as the depth of their lowest common ancestor allows: six for siblings (L1, L2, and L3 shared properties), four for entities sharing only an L2 ancestor, and two (the root-level properties) otherwise.

On top of the shared properties, one entity receives the probe property $P$, and its sibling is the target entity, for which $P$ is held out. Each of the six entities outside the target's L3 category receives its own control property, giving $|C| = 6$. Induction across categories, as measured in \autoref{subsec:evaluation}, is the probability the model assigns to a control property for the target entity, averaged over all control properties and prompt templates (\autoref{eq:induction_advantage}).

Each random seed regenerates the entire pseudoword vocabulary, both the entity names and the properties, and reassigns them to positions in the taxonomic tree. Running each experiment across three seeds therefore evaluates property induction with three different vocabularies and three different assignments to the tree, which checks that our results do not depend on any particular set of pseudowords.

\subsection{Constraining the completion space}
\label{app:completion_space}
 
The three quantities evaluated in the main text (memorization, induction within category, and induction across categories) are defined over property completions. However, comparing whole property phrases directly turns out to be unreliable. Each property consists of a verb and an adverb. Rather than sharing nothing between categories, we share the property verb across categories and compare only the probabilities of the token-length-matched adverbs that follow it. For example, instead of comparing the probability of \emph{gruffres branely} in \emph{hodger gruffres branely} against that of \emph{crambles peedly} in \emph{hodger crambles peedly}, we share the root verb \emph{gruffre} between categories and compare the completions \emph{branely} and \emph{peedly} after the common prefix \emph{hodger gruffres}.

This design serves two purposes. First, it constrains the space of plausible continuations, which sharpens the contrast between within-category and across-category induction, because both candidates are grammatically and distributionally supported at the position being scored. Second, it removes residual associations for entity names that resemble real words. For instance, \emph{cokol}, produced by the Wuggy generator, is close to the Czech word for falcon, \emph{sokol}. In our initial experiments, some entity names produced much higher entropy in the next-token distribution, which made results unstable across random seeds. Prefilling the property verb largely resolved this issue.

\subsection{Tokenization}
\label{app:tokenization}
Entity names occupy two or three tokens under both tokenizers, and no two entity names share the same last token. Probe and control adverbs are exactly token-length-matched, and no two adverbs in a taxonomy share their first token. The comparison between candidate completions is therefore decided at the first scored position rather than accumulating over a shared prefix, and no candidate is favored by being shorter: the induction advantage in \autoref{eq:induction_advantage} compares sums of log-probabilities over completions of equal length.
 
\subsection{Training and evaluation templates}
\label{app:templates}
 
\begin{templatebox}[label={box:train_templates}, float=t]{%
Training templates}
 
\textbf{\footnotesize Direct, present tense.} {\footnotesize The property is
realized as a finite verb phrase.}
\begin{tmpllist}
    \item \slot{entity} \slot{property}.
    \item It is known that \slot{entity} \slot{property}.
    \item Research confirms that \slot{entity} \slot{property}.
    \item Observations show that \slot{entity} \slot{property}.
    \item According to records, \slot{entity} \slot{property}.
    \item One key characteristic: \slot{entity} \slot{property}.
    \item Studies indicate that \slot{entity} \slot{property}.
    \item It has been documented that \slot{entity} \slot{property}.
    \item By all accounts, \slot{entity} \slot{property}.
    \item As widely reported, \slot{entity} \slot{property}.
\end{tmpllist}
 
\smallskip
\textbf{\footnotesize Indirect.} {\footnotesize The property is embedded under an
infinitive or raising construction and appears in its bare form.}
\begin{tmpllist}
    \item Based on extensive observation, \slot{entity} has been found to \slot{property}.
    \item Experts have long noted that \slot{entity} tends to \slot{property}.
    \item When studied closely, \slot{entity} was observed to \slot{property}.
    \item Historical records indicate that \slot{entity} is known to \slot{property}.
    \item In documented cases, \slot{entity} has been shown to \slot{property}.
    \item Careful analysis has shown that \slot{entity} appears to \slot{property}.
    \item Those familiar with \slot{entity} know it is prone to \slot{property}.
    \item It is not uncommon for \slot{entity} to \slot{property}.
    \item On closer inspection, \slot{entity} turns out to \slot{property}.
    \item Researchers have come to expect \slot{entity} to \slot{property}.
\end{tmpllist}
 
\smallskip
\textbf{\footnotesize Direct, past tense.} {\footnotesize The property is realized
in the past tense.}
\begin{tmpllist}
    \item In all previous iterations, \slot{entity} \slot{property}.
    \item Initial tests confirmed that \slot{entity} \slot{property}.
    \item Historically, it was found that \slot{entity} \slot{property}.
    \item According to the ancient logs, \slot{entity} \slot{property}.
    \item It was widely believed that \slot{entity} \slot{property}.
    \item Every time the team checked, \slot{entity} \slot{property}.
    \item The final report stated that \slot{entity} \slot{property}.
    \item Until the recent update, \slot{entity} \slot{property}.
    \item In past surveys, \slot{entity} \slot{property}.
    \item Observers were surprised when \slot{entity} \slot{property}.
\end{tmpllist}
\end{templatebox}
 
\begin{templatebox}[label={box:eval_templates}, float=t]{%
Evaluation templates}
 
\textbf{\footnotesize Direct, present tense.}
\begin{tmpllist}
    \item \slot{entity} \slot{property}.
    \item It is known that \slot{entity} \slot{property}.
    \item A defining trait is that \slot{entity} \slot{property}.
    \item It is a well-established fact that \slot{entity} \slot{property}.
\end{tmpllist}
 
\smallskip
\textbf{\footnotesize Indirect.} {\footnotesize These end in an infinitival marker or modal auxiliary and take the bare property form.}
\begin{tmpllist}
    \item \slot{entity} often tends to \slot{property}.
    \item Under typical conditions, \slot{entity} will \slot{property}.
    \item \slot{entity} has been observed to \slot{property}.
    \item Historical records show that \slot{entity} tends to \slot{property}.
\end{tmpllist}
 
\smallskip
\textbf{\footnotesize Direct, past tense.}
\begin{tmpllist}
    \item In earlier observations, \slot{entity} \slot{property}.
    \item Back then, \slot{entity} \slot{property}.
    \item It was found that \slot{entity} \slot{property}.
    \item In the past, \slot{entity} \slot{property}.
\end{tmpllist}
\end{templatebox}
 
Using a single fixed phrasing would let a model solve the task by matching one surface pattern, so we vary the sentence template while holding the underlying statements constant. Each training sentence is drawn uniformly from the pool of 30 templates in Box~\ref{box:train_templates}. The pool is split into three groups that differ in how the property is realized. Direct present-tense frames take the property as a finite verb phrase, as in \emph{hodger gruffres branely}. Indirect frames embed the property under an infinitive or raising construction, so the property appears in its bare form, as in \emph{pirer tends to gruffre branely}. Past-tense frames take the property in the past, as in \emph{colel gruffred branely}. We create these inflected forms (\eg, \emph{gruffres}, \emph{gruffre}, and \emph{gruffred}) manually from the Wuggy output.
 
At evaluation, we score property completions against the 12 templates in Box~\ref{box:eval_templates}. Of these, 2 also appear in the training pool and the remaining 10 are never seen during training. This held-out majority rules out the most immediate alternative explanation for our results, namely that the model has learned to complete a specific surface pattern rather than to project a property onto an entity. The evaluation templates follow the same three groups as the training pool, so the evaluation covers the same range of constructions the model was trained on, and averaging over them prevents any single phrasing from driving the measurement. Templates ending in an infinitival marker or modal auxiliary, such as \emph{often tends to} or \emph{will}, require the bare property form, and the completion is scored accordingly.

\subsection{Matching Affinity Clustering}
\label{app:clustering}
 
Matching Affinity Clustering (MAC) \citep{hajiaghayi2021improved} starts with all entities as singleton clusters. In each round, it partitions the current clusters into disjoint pairs by a minimum-weight perfect matching on the complete graph whose edge weights are the average-linkage cosine distances between clusters, then merges every matched pair.
 
Because each round matches every cluster exactly once, the number of clusters halves per round and the procedure yields a balanced binary tree in $\lceil \log_2 N \rceil$ rounds (for $N$ a power of two), matching the shape of our ground-truth taxonomy. In contrast, standard agglomerative clustering performs $N-1$ sequential merges, each joining the closest available pair, and generally produces unbalanced trees, which is why we do not use it in this work.

\subsection{Details of the taxonomy-conflict experiment}
\label{app:conflict_setup}
 
\begin{figure}[t]
    \centering
    \includegraphics[width=\linewidth]{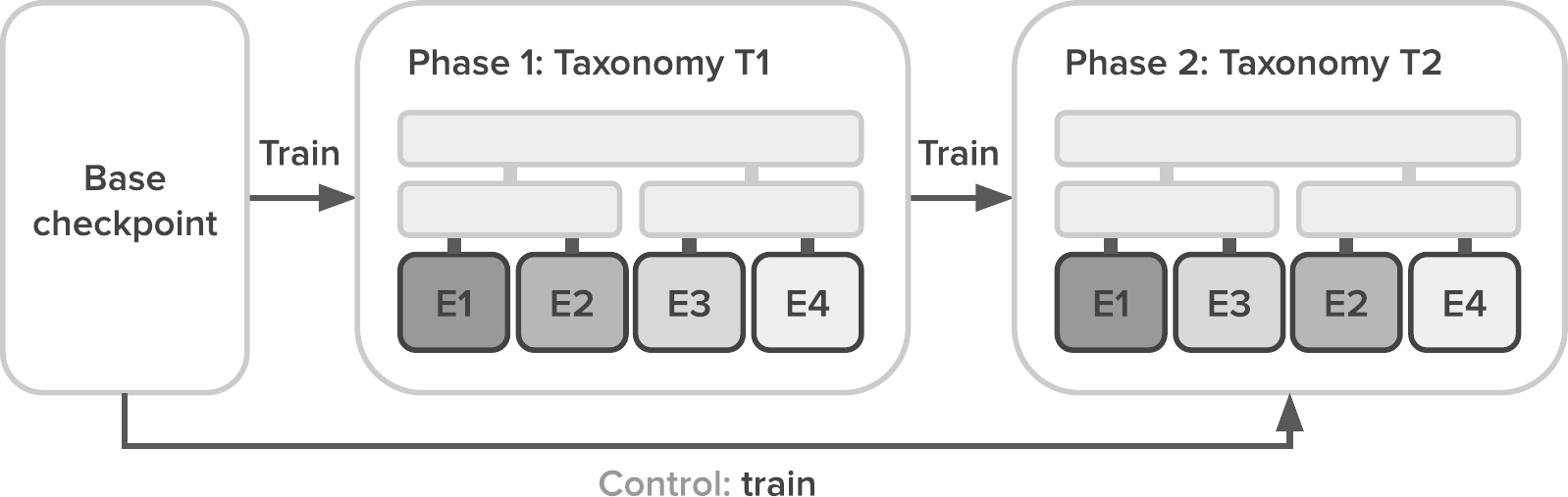}
    \caption{Schematic of the experiment on interference across taxonomies. Models are first trained on taxonomy $T_1$ and then on a conflicting taxonomy $T_2$ that reuses the same entity names but regroups them. \emph{Control} denotes standard training without taxonomy conflict (Taxonomy conflict: \emph{No}).}
    \label{fig:conflict_setup}
\end{figure}

\autoref{fig:conflict_setup} shows the two-phase design used in \autoref{sec:flexibility}. Starting from the same base checkpoint (step 50k of Pythia 160M and step 800k of OLMo 2 1B, as in \autoref{subsec:models}), models are first trained on taxonomy $T_1$ and then on a conflicting taxonomy $T_2$ that reuses the same entity names but regroups them into different categories, so that entities sharing a category under $T_1$ are separated under $T_2$. $T_2$ uses new shared properties, and the probe and control properties are reassigned so that the target entity of $P$ follows the $T_2$ grouping.

The control condition branches from the same base checkpoint and trains on $T_2$ directly, without the $T_1$ phase, using the same number of $T_2$ injection steps. The two conditions therefore differ only in the preceding $T_1$ phase, so any difference in how well $T_2$ is acquired can be attributed to interference from $T_1$ rather than to properties of $T_2$ itself. Both phases use the injection procedure and hyperparameters described in \autoref{app:hyperparams},
and at the start of $T_2$ training, both conditions use the optimizer state of the base checkpoint. The induction advantage is then measured with respect to both the $T_1$ and $T_2$ groupings throughout training. The values in \autoref{fig:conflict}~(right) are averaged over the last 1{,}000 steps of $T_2$ injection and over three random seeds.

\subsection{Presentation details}
\label{app:presentation}

All figures that plot memorization and induction within and across categories over training (\eg, \autoref{fig:main} and \autoref{fig:forgetting}) smooth the curves with a moving average over a window of 10 evaluation points.

In the schematics of the in-context experiments (\autoref{fig:incontext}) and of the taxonomy-conflict experiment (\autoref{fig:conflict_setup}), entities are labeled E1, E2, etc.\ for readability only; as in all other experiments, the actual runs use pseudoword entity names.

\section{Supporting results}
\label{app:results}

\subsection{Representational similarity by shared depth}
\label{app:repsim}
 
\begin{figure}[t]
    \centering
    \includegraphics[width=\linewidth]{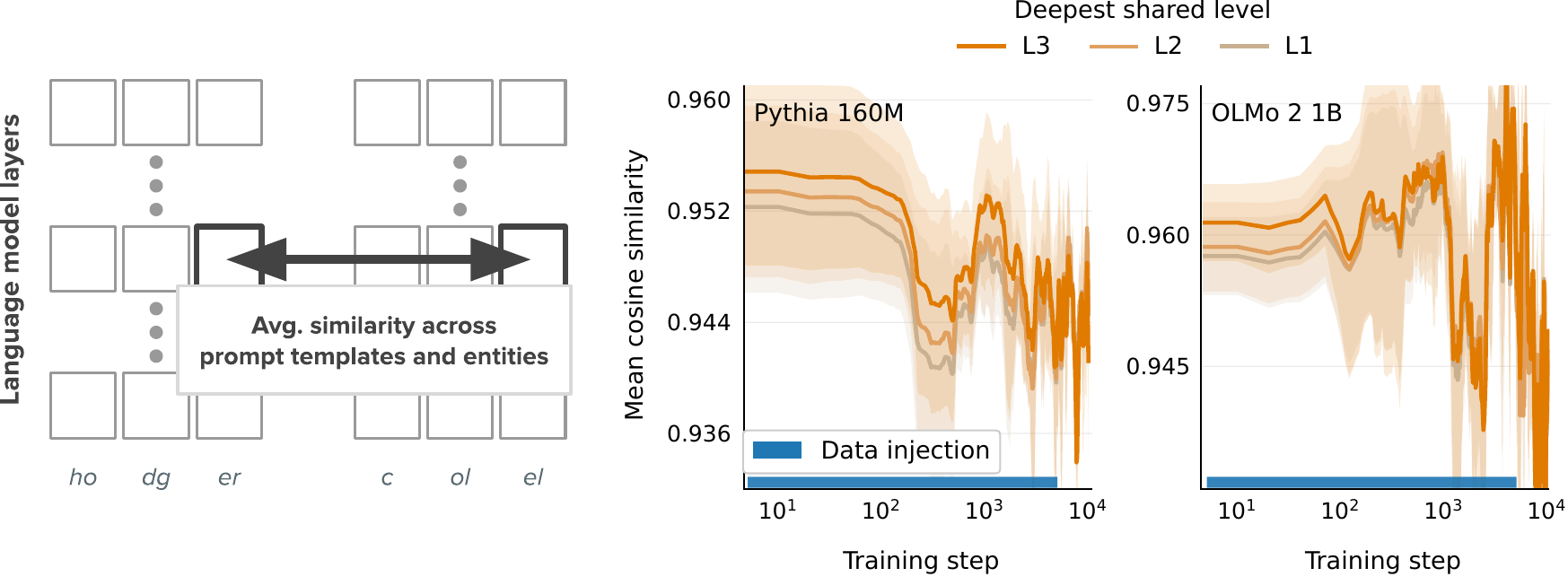}
    \caption{Mean pairwise cosine similarity between entity representations over continued pre-training, grouped by the deepest level of the ground-truth taxonomy that the two entities share. Similarity increases monotonically with shared depth (L1 $<$ L2 $<$ L3). Shaded regions indicate the standard error of the mean across three random seeds.}
    \label{fig:repsim}
\end{figure}
 
Before testing whether the full tree can be recovered, we check the weaker prediction of \citet{saxe2019}: that two entities sharing a deeper node should become more similar in representation than two sharing only a shallow one. We take last-token entity representations from a mid-to-late layer (layer 8 of 12 for Pythia~160M, layer 8 of 16 for OLMo~2~1B) and compute the mean pairwise cosine similarity within and between categories, averaged over all evaluation templates. Because a representation read at a token position reflects the surrounding context as well as the concept, we average over templates to keep the comparison about the entity rather than about the phrasing that contained it~\citep{park2025geometry}.
 
\autoref{fig:repsim} shows that average similarity increases monotonically with the deepest shared level of the ground-truth taxonomy (L1 $<$ L2 $<$ L3). The representation space is therefore reorganizing to match a structure that appears nowhere in the training text as an explicit statement.

\subsection{Recovery of the taxonomic tree}
\label{app:treerecovery}
 
\begin{figure}[t]
    \centering
    \includegraphics[width=\linewidth]{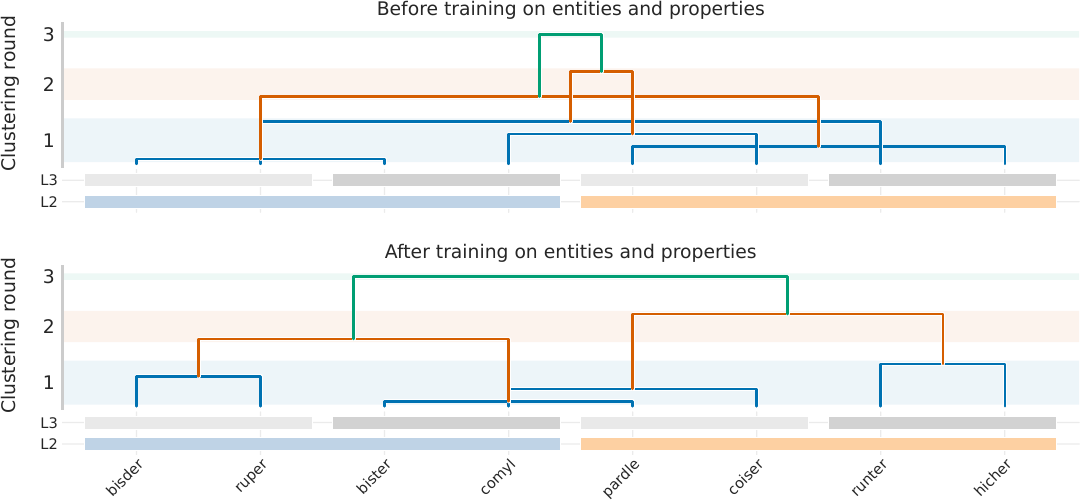}
    \vspace{-20pt}
    \caption{Hierarchical recovery of the taxonomy from entity representations of Pythia~160M, before (top) and after (bottom) continued pre-training on entities and properties. Bars below each dendrogram mark the ground-truth L3 and L2 categories.}
    \label{fig:cluster_pythia}
    \vspace{-4pt}
\end{figure}
 
\autoref{fig:cluster_pythia} shows the dendrograms recovered from Pythia~160M representations, complementing the OLMo~2~1B result in the main text.

\subsection{Forgetting after injection stops}
\label{app:forgetting}
 
\begin{figure}[t]
    \centering
    \includegraphics[width=\linewidth]{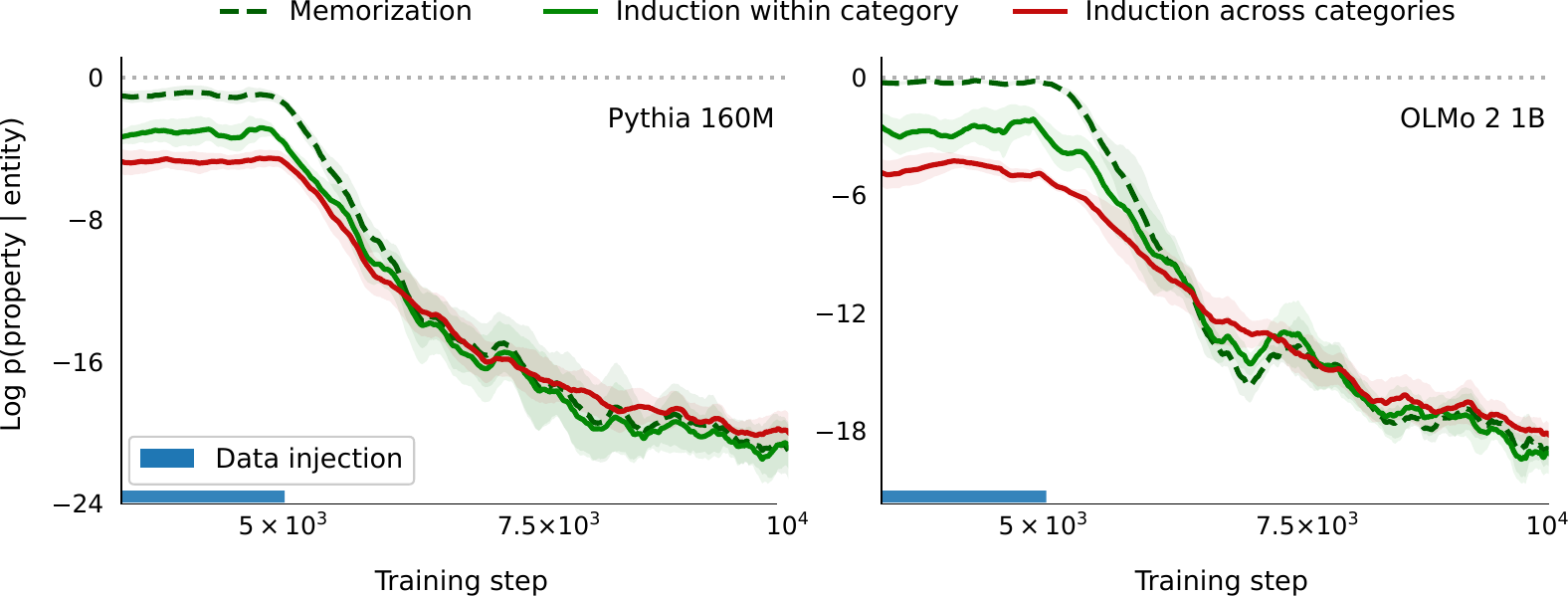}
    \vspace{-18pt}
    
    \caption{$\text{Log}\ p(\text{property} \mid \text{entity})$ over the last 6,000 continued pre-training steps of Pythia~160M (left) and OLMo~2~1B (right), after synthetic data injection stops at step 5,000. Shaded regions indicate the standard error of the mean across three random seeds.}
    \label{fig:forgetting}
    \vspace{-2pt}
\end{figure}
 
Correct induction grows over continued pre-training, so more exposure to the taxonomy yields stronger taxonomic generalization rather than only more memorization (\autoref{fig:main}). This generalization does not, however, persist on its own. When we stop injecting the synthetic data at step 5,000 and continue training on the background corpus alone, both memorization and correct induction decay within a few hundred steps (\autoref{fig:forgetting}). Taxonomic structure is therefore maintained only while the data continues to support it. \citet{chang2024knowledge} report comparable forgetting for individual injected facts during pre-training; our result shows that the taxonomic organization supporting their generalization is no more durable than the facts themselves.

\subsection{Memorization speed versus induction across model scale}
\label{app:scale_speed}
 
\begin{figure}[t!]
    \centering
    \includegraphics[width=\linewidth]{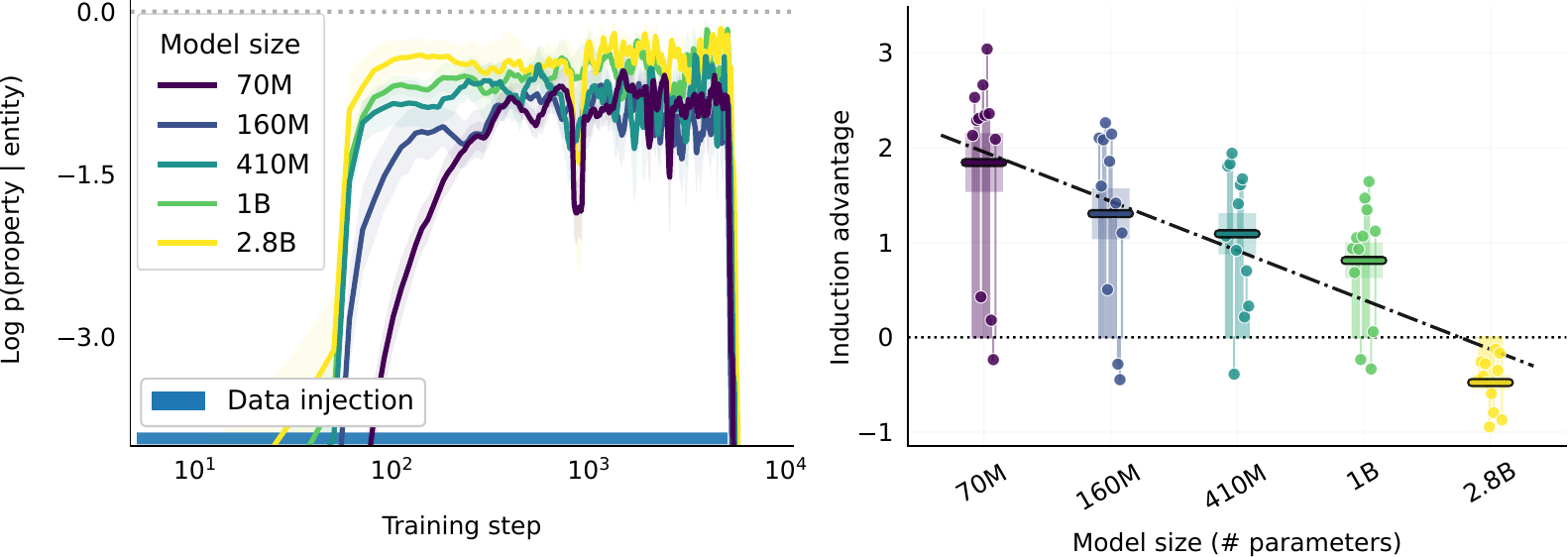}
    \caption{Memorization over training (left) and early induction advantage, averaged over steps 0--1{,}000 (right), versus model size for the Pythia family. Scale increases the speed of memorization while reducing correct inductive generalization. Circles indicate induction advantage for individual evaluation templates, while horizontal lines show the average. All reported values are averages across three random seeds, and the shaded regions show the standard error of the mean across those seeds.}
    \label{fig:scale_speed}
\end{figure}
 
To support the interpretation that, at a fixed data budget, scale buys memorization at the expense of correct inductive generalization, we compare the speed at which each is acquired. We run this analysis on the Pythia family, which provides a denser span of model scales. \autoref{fig:scale_speed} plots memorization over training for each model size (left) alongside the induction advantage averaged over the first 1,000 continued pre-training steps (right). The ordering of the curves on the left confirms that larger models memorize faster, while the downward trend on the right shows that they generalize along the taxonomy more slowly.

\subsection{The effect of batch size}
\label{app:batch_size}
 
\begin{figure}[t]
    \centering
    \includegraphics[width=0.82\linewidth]{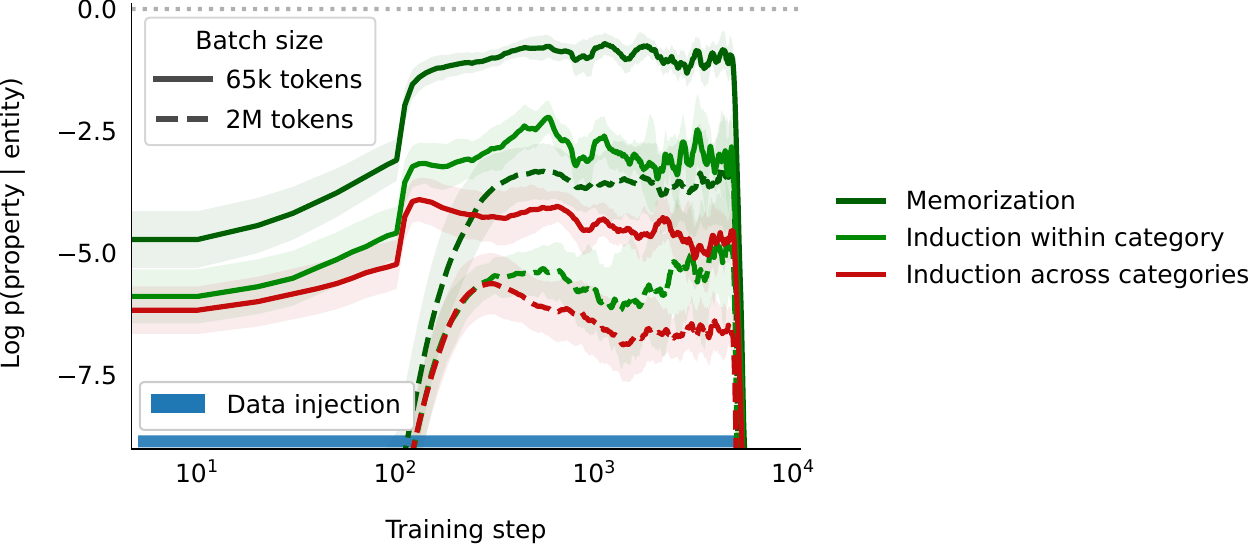}
    \caption{Comparison for Pythia~160M between the batch size used in all our experiments (32 sequences; 65k tokens) and the batch size of the original Pythia training run \citep{biderman2023pythia} (1{,}024 sequences; 2M tokens). Shaded regions indicate the standard error of the mean across three random seeds.}
    \label{fig:batch_size}
\end{figure}
 
All experiments in the paper use a batch size of 32 sequences, or roughly 65k tokens for Pythia and 130k tokens for OLMo~2. This is much smaller than the 1,024 sequences (2M tokens) and 512 sequences (2M tokens) used in the original Pythia and OLMo~2 runs \citep{biderman2023pythia,olmo2}. We chose the smaller batch size because of the available compute budget: continued pre-training has to be repeated across model sizes, checkpoints, seeds, and ablations, and the full batch size would have made that grid unaffordable. If our results held only at a batch size chosen for convenience, however, they would say more about our training setup than about LMs. We therefore repeat continued pre-training of Pythia~160M at the original batch size of 1,024 sequences, holding everything else fixed, including the taxonomy, the number of injected sequences per batch, the evaluation templates, and the seeds.
 
As \autoref{fig:batch_size} shows, the ordering of memorization above within-category induction above across-category induction is preserved at the larger batch size, so the qualitative finding does not depend on our choice. The differences are in the overall magnitude of all three quantities and in the speed at which correct induction is acquired: larger batches lead to slower differentiation between within- and across-category induction and to lower probabilities assigned to the synthetic properties. This follows from the ratio of synthetic entity-property data to the background corpus falling to $\approx$0.06--0.13\% after padding, which lowers the influence of the synthetic sequences on each training step.
 
What matters for the rest of the paper is which parts of the result are robust to this choice. The ordering of the three curves is the claim that our paper relies on, and it is the part that needs to be independent of batch size. Absolute values of the induction advantage are more fragile, which is why we compare them only within a fixed batch size and never across the two settings.

\end{document}